\documentclass[12pt]{article}

\usepackage{graphicx}
\usepackage{multicol}
\usepackage{amsmath}
\usepackage{amssymb}

\usepackage{geometry}
\usepackage{color}
\usepackage{bbold}
\usepackage{lscape}
\usepackage{setspace}
\usepackage{array}
\usepackage{multirow}
\usepackage{booktabs}
\usepackage{geometry}
\usepackage[font=small,labelfont=bf]{caption}
\usepackage{subfig}
\usepackage{fancybox}
\usepackage{enumitem}
\usepackage{tabularx}
\usepackage{float}
\usepackage{ifthen}
\usepackage{amsfonts}
\usepackage{amsthm}
\usepackage{eurosym}
\usepackage{setspace}
\usepackage{array}
\usepackage{multirow}
\usepackage{hyperref}
\usepackage{footmisc}

\usepackage{dsfont}
\usepackage[table]{xcolor}
\usepackage{lscape}
\usepackage{savefnmark}

\newcommand{\blanco}[1]{}

\makeatletter
\def\maxwidth{ %
  \ifdim\Gin@nat@width>\linewidth
    \linewidth
  \else
    \Gin@nat@width
  \fi
}
\makeatother

\definecolor{fgcolor}{rgb}{0.345, 0.345, 0.345}

\usepackage{framed}
\makeatletter
 {\par\unskip\endMakeFramed%
 \at@end@of@kframe}
\makeatother

\definecolor{shadecolor}{rgb}{.97, .97, .97}
\definecolor{messagecolor}{rgb}{0, 0, 0}
\definecolor{warningcolor}{rgb}{1, 0, 1}
\definecolor{errorcolor}{rgb}{1, 0, 0}

\usepackage{alltt}

\usepackage{mathptmx}

\usepackage{graphics}

\usepackage{natbib} 

\usepackage{tikz}
\usetikzlibrary{trees}
\usepackage{colortbl}
\usepackage{color}
\usepackage{booktabs}
\usepackage{hyperref}
\usepackage{amsmath}
\usepackage{amssymb}
\usepackage{float}
\usepackage{subfig}

\definecolor{lightgray}{rgb}{0.75, 0.75, 0.75}

\begin{document}

\title{Hybrid Machine Learning Forecasts for the FIFA Women's World Cup 2019}

\author{Andreas Groll
\thanks{Statistics Faculty, TU Dortmund University, Vogelpothsweg 87, 44227 Dortmund, Germany, \emph{groll@statistik.tu-dortmund.de}}
\and Christophe Ley
\thanks{Faculty of Sciences, Department of Applied Mathematics, Computer Science and Statistics, Ghent University, Krijgslaan 281, 9000 Gent,  Belgium, \emph{Christophe.Ley@UGent.be}}
\and Gunther Schauberger
\thanks{Chair of Epidemiology, Department of Sport and Health Sciences, Technical University of Munich, \emph{g.schauberger@tum.de}}
\and Hans Van Eetvelde
\thanks{Faculty of Sciences, Department of Applied Mathematics, Computer Science and Statistics, Ghent University, Krijgslaan 281, 9000 Gent,  Belgium, \emph{hans.vaneetvelde@ugent.be}}
\and Achim Zeileis
\thanks{Department of Statistics, Universit\"at Innsbruck, Austria, \emph{Achim.Zeileis@uibk.ac.at}}
}

\maketitle

\thispagestyle{empty}

\setlength{\parindent}{0pt}

\setlength{\columnsep}{15pt}

\textbf{Abstract}
{In this work, we combine two different ranking methods together with several other predictors
in a joint random forest approach for the scores of soccer matches.
The first ranking method is based on the bookmaker consensus, 
the second ranking method estimates adequate ability parameters that reflect the current strength of the teams best. 
The proposed combined approach is then applied to the data from 
the two previous FIFA Women's World Cups 2011 and 2015.
Finally, based on the resulting estimates, the FIFA Women's World Cup 2019 is simulated 
repeatedly and winning probabilities are obtained for all teams. The model clearly favors 
the defending champion USA  before the host France. 
}\bigskip


\section{Introduction}

While the research on adequate statistical models to predict the outcome of (men's) international soccer tournaments, 
such as European championships (EUROs) or FIFA World Cups, has substantially advanced in recent years, 
to the best of our knowledge there exists no significant scientific literature on modeling women's soccer.
In this work, we propose a combined ranking-based machine learning approach 
that is then used to forecast the FIFA Women's World Cup 2019.

A model class frequently used to model soccer results is the class of Poisson regression models. These 
directly model the number of goals scored by both competing teams in the single soccer matches.
Let $X_{ij}$ and $Y_{ij}$ denote the goals of the first and second team, respectively, in a match between teams $i$ and $j$, where $i,j\in\{1,\ldots,n\}$  and let $n$ denote the total number of teams in the regarded set of matches. 
One assumes $X_{ij}\sim Po(\lambda_{ij})$ and $Y_{ij}\sim Po(\mu_{ij})$, with intensity parameters 
$\lambda_{ij}$ and $\mu_{ij}$ (reflecting the expected numbers of goals). For these intensity parameters several modeling strategies exist, which incorporate playing abilities or covariates of the competing teams in different ways.

In the simplest case, the Poisson distributions are treated as (conditionally) independent, 
conditional on the teams' abilities or covariates. For example, \citet{Dyte:2000} applied this 
model to FIFA World Cup data, with Poisson intensities that depend on the FIFA ranks of 
both competing teams. \citet{GroAbe:2013} and \citet{GroSchTut:2015} 
considered a large set of potential predictors for EURO and World Cup data, 
respectively, and used $L_1$-penalized approaches 
to detect sparse sets of relevant covariates. Based on these, they calculated predictions for the EURO 
2012 and FIFA World Cup 2014 tournaments. Their findings show that, 
when many covariates are regarded and/or the predictive power of the single 
predictors is not clear in advance, regularization can be beneficial. 

These approaches can be generalized in different ways to allow for dependent scores. 
For example, \citet{DixCol:97} identified a (slightly negative) correlation between the scores
and introduced an additional dependence parameter. \citet{KarNtz:2003} and \citet{GrollEtAl2018} 
model the scores of both teams by a bivariate Poisson distribution, which is able to account for (positive) 
dependencies between the scores. If also negative dependencies should be accounted for, 
copula-based models can be used (see, e.g., \citealp{McHaSca:2007}, \citealp{McHaSca:2011} or \citealp{boshnakov2017}).

Closely related to the covariate-based Poisson regression models are Poisson-based 
ranking methods for soccer teams. On basis of a (typically large) set of matches, ability 
parameters reflecting the current strength of the teams can be estimated by means of maximum likelihood. 
An overview of the most frequently used Poisson-based ranking methods can be found in \citet{LeyWieEet2018}.

An alternative ranking approach that is solely based on bookmakers' odds was proposed by  
\citet{Leit:2010a}. They calculate winning probabilities for each team by aggregating winning
odds from several online bookmakers. Based on these
winning probabilities, by inverse tournament simulation
team-specific {\it bookmaker consensus abilities} can be computed by paired comparison
models, automatically stripping the effects of the tournament
draw. Next, pairwise probabilities for each
possible game at the corresponding tournament can be predicted
and, finally, the whole tournament can be simulated. 

A fundamentally different modeling approach is based on a
random forest -- a popular ensemble learning method for classification and regression \citep{Breiman:2001}, 
which originates from the machine learning and data mining community. 
Firstly, a multitude of so-called decision trees (\citealp{qui:1986}; \citealp{BreiFrieOls:84})
is constructed on different training data sets, which are resampled from the original dataset. The predictions from 
the individual trees are then aggregated, either by taking the mode of the predicted classes 
(in classification) or by averaging the predicted values (in regression). Random forests reduce 
the tendency of overfitting and the variance compared to regular decision trees, and are a common 
powerful tool for prediction. To investigate the predictive potential of random forests, \citet{SchauGroll2018} 
compared different types of random forests on data containing all matches of the FIFA 
World Cups 2002--2014 with conventional regression methods for count data, such as the 
Poisson models from above. The random forests provided very satisfactory results 
and generally outperformed the regression approaches. 
\citet{GroEtAl:WM2018b} showed on the same FIFA 
World Cup data that the predictive performance of random forests could 
be further improved by combining it with the Poisson ranking methods, leading to what they
call a \emph{hybrid random forest model}.

In the present work, we carry this strategy forward and combine the 
random forest with both the Poisson ranking methods as well as the bookmaker consensus abilities from \citet{Leit:2010a}.
So in a sense, this results in a {\it doubly-hybrid} or {\it combined ranking-based} random forest. 
The model is fitted to all matches of the FIFA Women's World Cups 2011 and 2015 and based on the resulting estimates, 
the FIFA Women's World Cup 2019 is then simulated 
100,000 times to determine winning probabilities for all 24 participating teams.

The rest of the manuscript is structured as follows. In Section~\ref{sec:data} we describe 
the three underlying data sets. The first covers all matches of the two preceding FIFA 
Women's World Cups 2011 and 2015 including covariate information, the second consists 
of the match results of all international matches played by all national teams during certain time periods 
and the third contains the winning odds from several bookmakers for the single World Cups regarded in this analysis.
Next, in Section~\ref{sec:methods} we briefly explain the basic idea of random forests and the two different
 ranking methods and, finally, how they can be combined to a hybrid random forest model.
In Section~\ref{sec:prediction}, we fit the hybrid random forest model to 
the data of the two World Cups 2011 and 2015. 
Based on the resulting estimates, the FIFA Women's World Cup 2019 is simulated 
repeatedly and winning probabilities for all teams are presented. 
Finally, we conclude in Section~\ref{sec:conclusion}.

\section{Data}\label{sec:data}

In this section, we briefly describe three fundamentally different types of data 
that can be used to model and predict international soccer tournaments such 
as the FIFA World Cup. The first type of data covers variables that characterize 
the participating teams of the single tournaments and connects them to the results 
of the matches that were played during these tournaments. 
The second type of data is simply based on the match results of all international 
matches played by all national teams during certain time periods. These data do not only cover 
the matches from the specific tournaments but also all qualifiers and friendly matches. 
The third type of data contains the winning odds from different bookmakers separately for
single World Cups.

\subsection{Covariate data}\label{sec:covariate}

The first type of data we describe covers all
matches of the two FIFA Women's World Cups 2011 and 2015 together with several
potential influence variables. Basically, we use a similar (but smaller\footnote{It turned out that, 
compared to men, for women's national teams covariates were generally more difficult to get or were simply not 
recorded at all, as for women's soccer data archives are less detailed and sometimes incomplete. For example, while 
for men's national coaches, their age, the duration of their tenure and their nationality 
could be obtained manually from the website of the German soccer magazine {\it kicker}, 
\url{http://kicker.de}, from \url{http://transfermarkt.de} and from \url{https://en.wikipedia.org}, this was not possible 
for women.}) set of covariates as introduced in \citet{GroSchTut:2015}. 
For each participating team, the covariates are observed either for the year of the respective World Cup 
(e.g.,\ GDP per capita) or shortly before the start of the World Cup (e.g.,\ average age), and, 
therefore, vary from one World Cup to another.

Several of the variables contain information about the recent performance 
and sportive success of national teams, as the current form of a national team is supposed to 
have an influence on the team's success in the upcoming tournament. One additional covariate in this regard,
which we will introduce later, is reflecting the national teams' current playing abilities and 
is related to the second type of data introduced in Section~\ref{sec:historic}. 
The estimates of these ability parameters are based on a separate Poisson ranking model, 
see Section~\ref{subsec:ranking} for details, and are denoted by {\it PoisAbil}. 
Another additional covariate, which is also introduced later, reflects 
the bookmaker consensus abilities (denoted by {\it OddsAbil}) from \citet{Leit:2010a} and is related to the third type of data 
introduced in Section~\ref{sec:bookmaker:data}. Details on this ranking method can be found in 
Section~\ref{subsec:consensus}.

Beside these sportive variables, also certain economic factors as well as variables describing the structure of a team's squad are collected. We shall now describe in more detail these variables.

\begin{description}
\item \textbf{Economic Factors:}

\begin{description}
\item[\it GDP per capita.] To account for the general 
	increase of the gross domestic product (GDP) during 2011--2015, a ratio of the GDP per capita of the respective country and the worldwide average GDP per capita is used (source: mostly \url{http://www.imf.org/external/pubs/ft/weo/2018/01/weodata/index.aspx}, but for England, Scotland, South and North Korea some additional research was necessary).
\item[\it Population.] The population size is 
	used in relation to the respective global population to account for the general  world population growth (source: \url{http://data.worldbank.org/indicator/SP.POP.TOTL}).
\end{description}\bigskip

\item \textbf{Sportive factors:}

\begin{description}
\item[\it FIFA rank.] The FIFA Coca-Cola Women's World Ranking is based on an Elo-type rating system, which was
originally developed by Dr.\ Arpad Elo to rate the playing abilities of chess players.  It aims at reflecting the current strength of a soccer team relative to its competitors (source: \url{https://de.fifa.com/fifa-world-ranking/ranking-table/women/}). 

\end{description}\bigskip

\item \textbf{Home advantage:}

\begin{description}
\item[\it Host.] A dummy variable 
indicating if a national team is a hosting country. 
\item[\it Continent.] A dummy variable indicating if a national team is from the same  continent as the host of the World Cup (including the host itself).
\item[\it Confederation.] This categorical variable comprises the teams' confederation with six possible values: Africa (CAF);
    Asia (AFC); Europe (UEFA); North, Central America and Caribbean (CONCACAF); Oceania (OFC); South America (CONMEBOL). 
\end{description}
\bigskip

\item \textbf{Factors describing the team's structure:}

	The following variables describe the structure of the teams. 
	They were observed with the 23-player-squad 
	nominated for the respective World Cup and were obtained manually 
	both from the website of the German soccer magazine {\it kicker}, \url{http://kicker.de}, and 
	from \url{http://transfermarkt.de}\footnote{Note that for the World Cup 2011 the size of the 
	national teams' squads was restricted to 21 players. Hence, all of the following factors that 
	add up players with a certain characteristic (namely all factors except for the {\it average age}) have been divided by the respective 
	squad size (i.e.\ 21 or 23) to make them comparable across tournaments.}.\medskip 
	
\begin{description}
\item[\it (Second) maximum number of teammates.] For each squad, both the maximum 
	and second maximum number of teammates playing together in the same national club are counted.
\item[\it Average age.] The average age of each squad is collected. 
\item[\it Number of Champions League players.] 	
As a measurement of the success of the players on club level, the number of 
	players in the semi finals (taking place only few weeks before the 
	respective World Cup) of the UEFA Champions 
	League (CL) is counted. 
\item[\it Number of Major League Soccer players.] 	
As the US Major League Soccer (MLS) is supposedly the best 
national soccer league on the globe in women's soccer,
for each national team the number of players in the MLS is counted. 
\item[\it Number of players abroad/Legionnaires.] For each squad, the number of players 
	playing in clubs abroad (in the season preceding the respective World Cup) is counted.
\end{description}
\end{description}

\noindent In addition, we include a dummy variable indicating whether a certain match is a group- or a knockout match.
The motivation for this is that soccer teams might change their playing style and be more cautious in knockout matches.
In total, together with the two ability variables from the two ranking methods this adds up to 15 variables which were collected separately for each World Cup and each participating team. As an illustration, Table~\ref{data1} shows the results (\ref{tab:results}) and (parts of) the covariates (\ref{tab:covar}) of the respective teams, exemplarily for the first four matches of the FIFA Women's World Cup 2011. We use this data excerpt to illustrate how the final data set is constructed.

	\begin{table}[h]
\small
\caption{\label{data1} Exemplary table showing the results of four matches and parts of the covariates of the involved teams.}
\centering
\subfloat[Table of results \label{tab:results}]{
\begin{tabular}{lcr}
  \hline
 &  &  \\ 
  \hline
NGA \includegraphics[width=0.4cm]{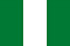} & 0:1 &  \includegraphics[width=0.4cm]{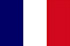} \;FRA\\
GER \includegraphics[width=0.4cm]{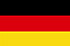} & 2:1 &  \includegraphics[width=0.4cm]{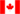} \;CAN\\
CAN\includegraphics[width=0.4cm]{CAN.png} & 0:4 &  \includegraphics[width=0.4cm]{FRA.png} \;FRA\\
GER \includegraphics[width=0.4cm]{GER.png} & 1:0 &  \includegraphics[width=0.4cm]{NGA.png} \;NGA\\
  \vdots & \vdots & \vdots  \\
  \hline
\end{tabular}}
\hspace*{0.8cm}
\subfloat[Table of covariates \label{tab:covar}]{
\begin{tabular}{llrrrrr}
  \hline
World Cup & Team &  PoisAbil & OddsAbil &   Age & \ldots \\ 
  \hline
2011 & France  & $1.69$  &  $0.02$ & $25.86$ & \ldots \\ 
2011 &  Germany & $2.35$  &  $1.25$ & $25.95$ & \ldots\\ 
2011 &  Nigeria & $1.39$  & $-0.47$ & $22.24$ & \ldots \\ 
2011 &  Canada & $1.82$  & $-0.17$ & $25.52$ & \ldots\\ 
  \vdots & \vdots & \vdots & \vdots & \vdots  &  $\ddots$ \\
   \hline
\end{tabular}
}
\end{table}

For the modeling techniques that we shall introduce in the following sections, all of the metric 
covariates are incorporated in the form of differences between the two competing teams. For example, the final 
variable {\it PoisAbil} will be the difference between the Poisson ranking abilities of both teams. The categorical variables {\it Host},
{\it Continent} and {\it Confederation}, however, are included as separate variables for both competing teams.
For the variable {\it Confederation}, for example, this results in two columns of the corresponding design matrix denoted by 
{\it Confed} and {\it Confed.Oppo}, where {\it Confed} is referring to the confederation of the first-named team
and {\it Confed.Oppo} to the one of its opponent. 

As we use the number of goals of each team directly as the response variable, each match 
corresponds to two different observations, one per team. For the covariates, we consider 
differences which are computed from the perspective of the first-named team. 
The dummy variable {\it groupstage} corresponds to a single column in the 
design matrix and is either zero or one for both rows corresponding to the same match.For illustration, 
the resulting final data structure for the exemplary matches from Table~\ref{data1} is displayed in Table~\ref{data2}.

\begin{table}[!h]
\small
\centering
\caption{Exemplary table illustrating the data structure.}\label{data2}
\begin{tabular}{rllrrrrr}
  \hline
Goals & Team & Opponent & Groupstage & PoisAbil & OddsAbil &   Age &  ... \\ 
  \hline
0 & Nigeria & France & 1 & $-0.49$  &  $-0.30$ & $-3.62$ &  ...  \\ 
  1 & France & Nigeria & 1 & $0.49$ &  $0.30$ & $3.62$ &  ...  \\ 
  2 & Germany & Canada & 1 & $1.42$  &  $0.53$ & $0.43$ &  ...  \\ 
  1 & Canada & Germany & 1 & $-1.42$  &  $-0.53$ & $-0.43$ &  ...  \\ 
    0 & Canada & France & 1 & $-0.18$  &  $0.13$ & $-0.33$ &  ... \\ 
  4 & France & Canada & 1 & $0.18$  &  $-0.13$ & $0.33$ &  ... \\ 
  1 & Germany & Nigeria & 1 & $1.73$  & $0.96$ & $3.71$ &  ...  \\ 
  1 & Nigeria & Germany & 1 & $-1.73$  & $-0.96$ & $-3.71$ &  ... \\ 
	 \vdots & \vdots & \vdots & \vdots & \vdots & \vdots & \vdots &  $\ddots$ \\
   \hline
\end{tabular}
\end{table}

\subsection{Historic match results}\label{sec:historic}

The data used for estimating the abilities of the teams consist of the results of every international game played in the last 8 years preceding the considered World Cup. Besides the number of goals, we also need the information of the venue of the game in order to correct for the home effect and the moment in time when a match was played. 
 The reason is that, in the ranking method described in Section~\ref{subsec:ranking}, 
 each match is assigned a weight depending on the time elapsed since the game took place. For example, Table~\ref{tab:historicdata}
  shows an excerpt of the historic match data used to obtain ability estimates for the teams at the FIFA Women's World Cup 2011. 
\begin{table}[ht]
\caption{Historical match result data used for estimating the abilities, 
exemplarily for the FIFA Women's World Cup 2011}
\label{tab:historicdata}
\centering
\begin{tabular}{lllcrr}
  \hline
  Date & Home team & Away team & Score & Country & Neutral \\ 
  \hline
 2011-05-19 & Iceland & Bulgaria &   6:0 &  Iceland & no  \\ 
 2011-03-09 & United States & Iceland &   4:2 &  Portugal & yes  \\ 
 2011-03-09 & Portugal & Finland &   2:1  & Portugal & no  \\ 
 2011-03-09 & Wales & China PR &   1:2 &  Portugal & yes   \\
 $\vdots$ & $\vdots$ & $\vdots$ & $\vdots$ & $\vdots$ & $\vdots$  \\
   \hline
\end{tabular}
\end{table}

\subsection{Bookmaker data}\label{sec:bookmaker:data}

The basis for the bookmaker consensus ranking model from \citet{Leit:2010a}, which is explained in more 
detail in Section~\ref{subsec:consensus}, are the winning odds of the bookmakers\footnote{The possibility of betting on the overall cup winner before the start of the tournament is quite novel. While for men, the German state
betting agency ODDSET offered the bet for the first time at the FIFA
World Cup 2002, for women we could not get any odds before the World Cup 2011.}. These are typically available 
already a few weeks before the tournament start. The popularity of this specific bet has substantially increased over time:
while for the World Cup 2011 we could only obtain the odds from the German state
betting agency ODDSET (upon request), for the World Cup 2015 we found already 
corresponding odds of three different bookmakers publicly available, see Table~\ref{tab:consensus}.
For the upcoming World Cup 2019 tournament we easily obtained the winning odds from 18 different bookmakers.

\begin{table}[H]
\centering
\caption{Winning odds of all 24 participating teams from several bookmakers, 
exemplarily for the FIFA Women's World Cup~2015}\label{tab:consensus}
\begin{tabular}{rrrr}
  \hline
 & MyTopSportsbooks & SportsInsights & BovadaSportsbook \\ 
  \hline
United States & 4.00 & 4.00 & 3.25 \\ 
  Germany & 4.50 & 4.35 & 4.50 \\ 
  France & 5.50 & 10.00 & 9.00 \\ 
  Japan & 9.00 & 9.00 & 8.00 \\ 
  Brazil & 9.00 & 8.50 & 8.00 \\ 
  Canada & 15.00 & 15.00 & 11.00 \\ 
  Sweden & 17.00 & 15.00 & 11.00 \\ 
  England & 21.00 & 21.00 & 21.00 \\ 
  Norway & 34.00 & 28.00 & 26.00 \\ 
  Australia & 51.00 & 51.00 & 41.00 \\ 
  China PR & 61.00 & 66.00 & 51.00 \\ 
  Spain & 67.00 & 66.00 & 41.00 \\ 
  Netherlands & 81.00 & 76.00 & 51.00 \\ 
  South Korea & 81.00 & 76.00 & 67.00 \\ 
  Switzerland & 101.00 & 86.00 & 67.00 \\ 
  New Zealand & 151.00 & 151.00 & 101.00 \\ 
  Nigeria & 201.00 & 201.00 & 151.00 \\ 
  Colombia & 251.00 & 251.00 & 151.00 \\ 
  Mexico & 251.00 & 251.00 & 126.00 \\ 
  Ecuador & 501.00 & 501.00 & 251.00 \\ 
  Cameroon & 501.00 & 501.00 & 301.00 \\ 
  Ivory Coast & 501.00 & 501.00 & 251.00 \\ 
  Costa Rica & 1001.00 & 1001.00 & 251.00 \\ 
  Thailand & 5001.00 & 5001.00 & 401.00 \\ 
   \hline
\end{tabular}
\end{table}

\section{A combined ranking-based random forest}\label{sec:methods}

In this section, we propose to use a hybrid random forest approach that combines the 
information from all three types of data bases introduced above. The proposed method combines 
a random forest for the covariate data with both the abilities estimated on the historic match results 
as used by the Poisson ranking methods and the abilities obtained from the bookmaker consensus approach. 
Before introducing the proposed hybrid method, we first separately present the 
basic ideas of the three model components.


\subsection{Random forests}\label{subsec:forest}

Random forests, originally proposed by \citet{Breiman:2001}, are an  
aggregation of a (large) number of classification or regression trees (CARTs). 
CARTs \citep{BreiFrieOls:84} repeatedly partition the predictor space mostly using 
binary splits. The goal of the partitioning process is to find partitions such that the 
respective response values are very homogeneous within a partition but very 
heterogeneous between partitions. CARTs can be used both for metric response 
(regression trees) and for nominal/ordinal responses (classification trees). 
For prediction, all response values within a partition are aggregated either 
by averaging (in regression trees) or simply by counting and using majority vote (in classification trees).
In this work, we use trees (and, accordingly, random forests) for the prediction of 
the number of goals a team scores in a match of a FIFA World Cup.

As already mentioned in the Introduction, random forests are the aggregation of a large number 
$B$ (e.g., $B=5000$) of trees, grown on $B$ bootstrap samples from the original data set. 
Combining many trees has the advantage
 that the resulting predictions inherit the feature of unbiasedness from the single trees 
 while reducing the variance of the predictions. For a short introduction to random forests 
and how they can specifically 
be used for soccer data, see \citet{GroEtAl:WM2018b}.

 

In \texttt{R} \citep{RDev:2018}, two slightly different variants of regression forests are 
available: the classical random forest algorithm proposed by \citet{Breiman:2001} 
from the \texttt{R}-package \texttt{ranger} \citep{ranger}, and a modification implemented 
in the function \texttt{cforest} from the \texttt{party} package\footnote{Here, the single trees are 
constructed following the principle of conditional inference trees as proposed in 
\citet{Hotetal:2006}. The main advantage of these conditional inference trees is 
that they avoid selection bias if covariates have different scales,
 e.g., numerical vs. categorical with many categories (see, for example, 
 \citealp{StrEtAl07}, and \citealp{Strobl-etal:2008}, for details). Conditional 
 forests share the feature of conditional inference trees of avoiding biased variable selection.}. 
%
In  \citet{SchauGroll2018} and \citet{GroEtAl:WM2018b}, the latter package
turned out to be superior and will be used in the following.

\subsection{Poisson ranking methods}\label{subsec:ranking}

In this section we describe how (based on historic match data, see Section~\ref{sec:historic}) Poisson models can be used to obtain rankings that reflect a team's current ability. We will restrict our attention to the best-performing model according to the comparison achieved in \cite{LeyWieEet2018}, namely the bivariate Poisson model. The main idea consists in assigning a strength parameter to every team and in estimating those parameters over a   period of $M$ matches via weighted maximum likelihood based on time depreciation.

The time decay function is defined as follows: a match  played $x_m$ days back gets a weight of 
\begin{equation*}\label{smoother}
w_{time,m}(x_m) = \left(\frac{1}{2}\right)^{\frac{x_m}{\mbox{\small Half period}}},
\end{equation*}
meaning that, for instance, a match played \emph{Half period} days ago only contributes half as much as a match played today. We stress that the \emph{Half period} refers to calendar days in a year, not match days. In the present case we use a Half period of 500 days based on an optimization procedure to determine which Half period led to the best prediction for women's soccer matches based on the average Rank Probability Score (RPS; \citealp{Gneitingetal:2007})

The bivariate Poisson ranking model is based on a proposal from \cite{KarNtz:2003} and can be described as follows. If we have $M$ matches featuring a total of $n$ teams, we write $Y_{ijm}$  the random variable \textit{number of goals scored by team $i$ against team $j$ ($i,j\in \{1,...,n\}$) in match $m$} (where $m \in \{1,...,M\}$). The joint probability function of the home and away score is then given by the bivariate Poisson probability mass function, 
\begin{eqnarray*} 
{\rm P}(Y_{ijm}=z, Y_{jim}=y) &=& \frac{\lambda_{ijm}^z \lambda_{jim}^y}{z!y!} \exp(-(\lambda_{ijm}+\lambda_{jim}+\lambda_{C}))\cdot\nonumber\\
&&\sum_{k=0}^{\min(z,y)} \binom{z}{k} \binom{y}{k}k!\left(\frac{\lambda_{C}}{\lambda_{ijm}\lambda_{jim}}\right)^k, 
\end{eqnarray*}
where $\lambda_{C}$ is a covariance parameter assumed to be constant over all matches and $\lambda_{ijm}$ is the expected number of goals for team $i$ against team $j$ in  match $m$, which we model as
\begin{eqnarray}
\label{independentpoisson}\log(\lambda_{ijm})&=&\beta_0 + (r_{i}-r_{j})+h\cdot \mathbb{I}(\mbox{team $i$ playing at home})\,,
\end{eqnarray}
where $\beta_0$ is a common intercept and $r_i$ and $r_j$ are the strength parameters of team~$i$ and team~$j$, respectively. Since the ratings are unique up to addition by a constant, we add the constraint that the sum of the ratings has to equal zero. The last term $h$ represents the home effect and is only added if team~$i$ plays at home. Note that we have the Independent Poisson model if $\lambda_C=0$. The overall  (weighted) likelihood function then reads
\begin{equation*}
L = \prod_{m=1}^{M}\left({\rm P}(Y_{ijm}=y_{ijm}, Y_{jim}={y_{jim}})\right)^{w_{time,m}}, 
\end{equation*}
where $y_{ijm}$ and $y_{jim}$ stand for the actual number of goals scored by teams $i$ and $j$ in match $m$. The values of the strength parameters $r_1,\ldots,r_n$, which allow ranking the different teams, are computed numerically as maximum likelihood estimates {on the basis of historic match data as described in Section~\ref{sec:historic}}.  These parameters also allow to predict future match outcomes thanks to the formula~\eqref{independentpoisson}.

\subsection{The bookmaker consensus ranking model}\label{subsec:consensus}

Prior to the tournament on 2019-06-03 we obtained long-term winning odds from 18~online bookmakers.
However, before these odds can be transformed to winning probabilities, the stake has to be accounted for
and the profit margin of the bookmaker (better known as the ``overround'') has to be removed
\citep[for further details see][]{ref:Henery:1999, ref:Forrest+Goddard+Simmons:2005}. 
Here, it is assumed that the quoted odds are derived from the underlying ``true'' odds as:
$\mbox{\it quoted odds} = \mbox{\it odds} \cdot \delta + 1$,
where $+ 1$ is the stake (which is to be paid back to the bookmakers' customers in case they win)
and $\delta < 1$ is the proportion of the bets that is actually paid out by the bookmakers.
The overround is the remaining proportion $1 - \delta$ and the main basis of the bookmakers'
profits (see also \citealp{ref:Wikipedia:2019} and the links therein).
Assuming that each bookmaker's $\delta$ is constant across the various teams in the tournament
\citep[see][for all details]{Leit:2010a}, we obtain overrounds for all
bookmakers with a median value of 24.8\%.

To aggregate the overround-adjusted odds across the 18~bookmakers, we transform them to
the log-odds (or logit) scale for averaging \citep[as in][]{Leit:2010a}.
The bookmaker consensus is computed as the mean winning log-odds for each team across bookmakers
and then transformed back to the winning probability scale.

In a second step the bookmakers' odds are employed to infer the contenders' relative abilities
(or strengths). To do so, an ``inverse'' tournament simulation based on team-specific
abilities is used. The idea is the following:
\begin{enumerate}
  \item If team abilities are available, pairwise winning probabilities can be derived
    for each possible match using the classical \cite{ref:Bradley+Terry:1952} model. This model is
    similar to the Elo rating \citep{ref:Elo:2008}, popular in sports, and computes the
    probability that a Team~$A$ beats a Team~$B$ by their associated abilities (or strengths):
    \[
    \mathrm{Pr}(A \mbox{ beats } B) = \frac{\mathit{ability}_A}{\mathit{ability}_A + \mathit{ability}_B}.
    \]
  \item Given these pairwise winning probabilities, the whole tournament can be easily
    simulated to see which team proceeds to which stage in the tournament and which
    team finally wins.
  \item Such a tournament simulation can then be run sufficiently often (here 1,000,000 times)
    to obtain relative frequencies for each team winning the tournament.
\end{enumerate}
Here, we use the iterative approach of \cite{Leit:2010a}
to find team abilities so that the resulting simulated winning probabilities
(from 1,000,000 runs) closely match the bookmaker consensus probabilities. This allows to
strip the effects of the tournament draw (with weaker/easier and stronger/more difficult
groups), yielding a log-ability measure (on the log-odds scale) for each team.

\subsection{The combined ranking-based random forest}

In order to link the information provided by the covariate data, the historic match data 
and the bookmakers' odds, we now combine the random forest approach from 
Section~\ref{subsec:forest} and the ranking methods from Section~\ref{subsec:ranking}
and Section~\ref{subsec:consensus}. We propose to use both ranking approaches 
to generate two new (highly informative) covariates that can be incorporated into the random forest model. 
For that purpose, for each World Cup we estimate the team abilities $r_i, i=1,\ldots,24$, of all $24$ 
participating teams shortly before the start of the respective tournament. For example, 
to obtain ability estimates for the 24 teams that participated in the World Cup 2011, 
the historic match data for a certain time period preceding the World Cup 2011 
(we chose to use 8 years, weighted by the described time 
depreciation effect) is used. This procedure gives us the estimates $\hat r_i$ as an additional covariate 
covering the current strength for all teams participating in a certain World Cup. Actually, this variable 
appears to be somewhat similar to the Elo ranking, but turns out to be much 
more informative, see Section~\ref{sec:fitforest}. 
Moreover, based on the winning odds provided by the bookmakers,
we also calculate the additional abilities $s_i, i=1,\ldots,24$, 
of all $24$ participating teams shortly before the start of the respective tournament
corresponding to the bookmaker consensus model. The corresponding estimates
 $\hat s_i$ again serve as another additional covariate. Also this variable turns out to be more important than the Elo ranking, see again Section~\ref{sec:fitforest}.

The newly generated variables can be added to the covariate data based on 
previous World Cups and a random forest can be fitted to these data. Based on 
this random forest, new matches (e.g., matches from an upcoming World Cup) can be predicted.
To predict a new observation, its covariate values are dropped down from each of the $B$ 
regression trees, resulting in $B$ distinct predictions. The average of those is then used as a 
point estimate of the expected numbers of goals conditioning on the covariate values. 
In order to be able to use these point estimates for the prediction of the 
outcome of single matches or a whole tournament, we follow \citet{GroEtAl:WM2018b} 
and treat the predicted expected value for the number of goals as 
an estimate for the intensity $\lambda$ of a Poisson distribution $Po(\lambda)$.
This way we can randomly draw results for single matches and compute 
probabilities for the match outcomes \textit{win}, \textit{draw} and \textit{loss} by 
using two independent Poisson distributions (conditional on the covariates) for both scores.

\section{Modeling the FIFA Women's World Cup 2019}\label{sec:prediction}
We now fit the proposed combined ranking-based random forest model 
to the data of the World Cups 2011 and 2015. 
Next, we calculate the Poisson ranking ability parameters 
based on historic match data over the 8 years preceding the World Cup 2019 as well as the bookmaker consensus abilities based on
the winning odds from 18 different bookmakers. 
Based on these ability estimates, the fitted random forest will be used to 
simulate the FIFA Women's World Cup 2019 tournament 
100,000 times to determine winning probabilities for all 24 participating teams.

\subsection{Fitting the combined ranking-based random forest to the data of the World Cups 2011 and 2015}\label{sec:fitforest}

\begin{figure}[!ht]
	\centering
		\includegraphics[width=.9\textwidth]{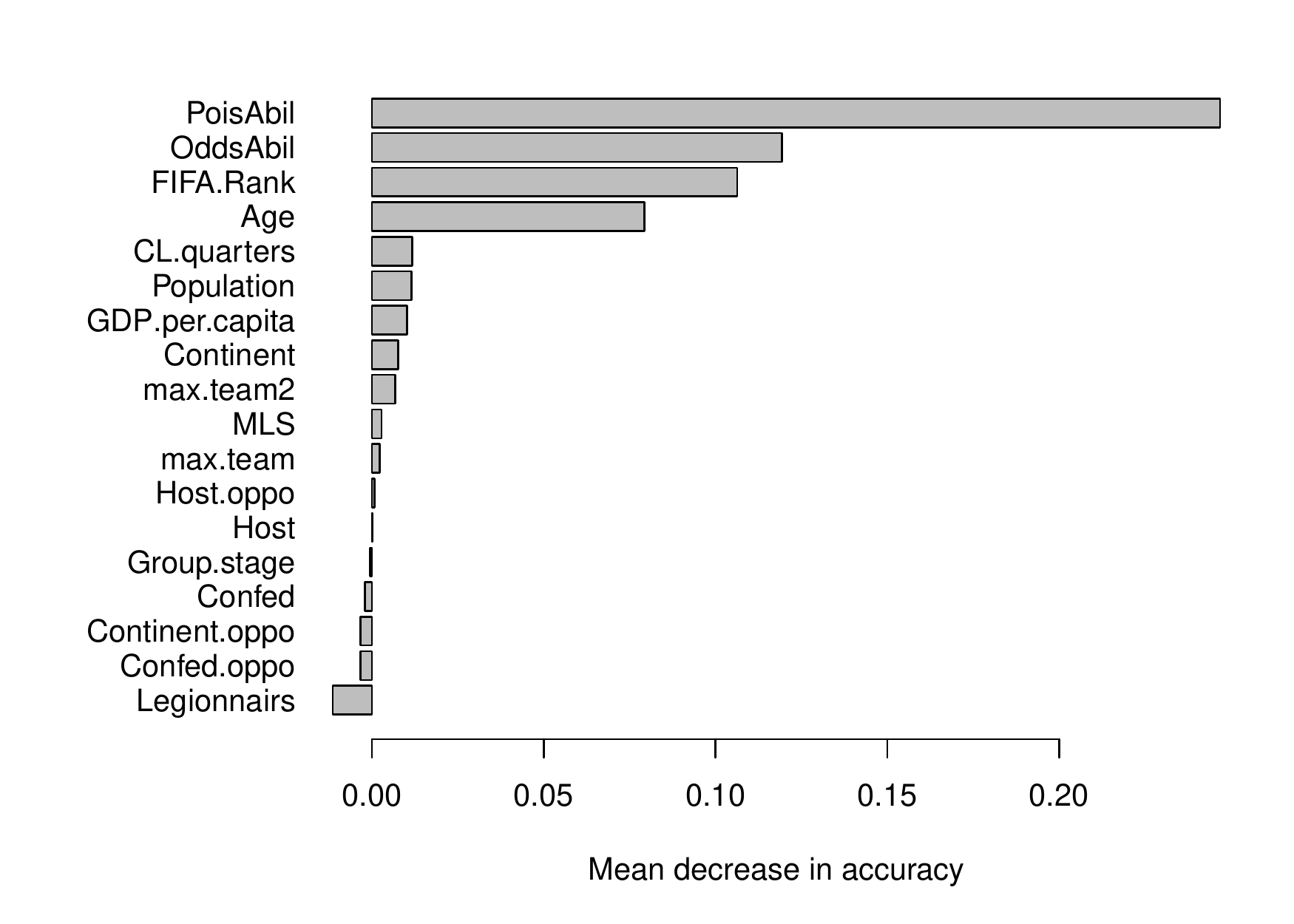}
	\caption{Bar plot displaying the variable importance in the hybrid random forest applied to FIFA World Cup data.}
	\label{var_imp}
\end{figure}

We fit the hybrid random forest approach with $B=5000$ single trees to the 
complete data set covering the two World Cups 2011 and 2015. 
The  best way to understand the role of the single predictor variables in a random forest
is the so-called variable importance, see \citet{Breiman:2001}. Typically, the variable 
importance of a predictor is measured by permuting each of the predictors 
separately in the out-of-bag observations of each tree. Out-of-bag observations 
are observations which are not part of the respective subsample or bootstrap 
sample that is used to fit a tree. Permuting a variable means that within the 
variable each value is randomly assigned to a location within the vector. 
If, for example, \emph{Age} is permuted, the average age of the German team
 in 2011 could be assigned to the average age of the US team in 2015. 
 When permuting variables randomly, they lose their information with respect 
 to the response variable (if they have any). Then, one measures the loss of prediction 
 accuracy compared to the case where the variable is not permuted. 
 Permuting variables with a high importance will lead to a higher loss of 
 prediction accuracy than permuting values with low importance. 
Figure~\ref{var_imp} shows bar plots of the variable importance values for all variables in the 
hybrid random forest applied to the data of the World Cups 2011 and 2015. 
Interestingly, the Poisson abilities are by far the most important predictor
in the random forest and carry clearly more information than all other predictors. 
But also the abilities from the bookmaker consensus approach seem to be 
slightly more informative compared to the \emph{FIFA rank}.
Even though the \emph{FIFA rank}, the teams' average \emph{age} or the number of \emph{CL~players}
also contain some information concerning the current strengths of the teams, 
it is definitely worth the effort to estimate such abilities in separate models. 
For a more detailed comparison of the team abilities and the \emph{FIFA rank},
see Table~\ref{tab_rank}.

%

\begin{table}[!h]
\small

\caption{\label{tab_rank} Ranking of the participants of the FIFA World Cup 2019 
according to estimated bookmaker consensus abilities (left; in logs), Poisson abilities (center; in logs) and FIFA ranking (right).}\vspace{0.4cm}
\centering
\begin{tabular}{l|ll|ll|ll}
\multicolumn{1}{c}{}&\multicolumn{2}{c}{\textbf{bookmaker consensus}} &\multicolumn{2}{c}{\textbf{Poisson}}  &\multicolumn{2}{c}{\textbf{FIFA}}\\
\multicolumn{1}{c}{}&\multicolumn{2}{c}{\textbf{abilities}} &\multicolumn{2}{c}{\textbf{abilities}}  &\multicolumn{2}{c}{\textbf{ranking}}\\
\toprule
1 & \includegraphics[width=0.4cm]{FRA.png} & France & \includegraphics[width=0.4cm]{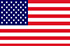} & United States & \includegraphics[width=0.4cm]{USA.png} & United States \\ 
  2 & \includegraphics[width=0.4cm]{USA.png} & United States & \includegraphics[width=0.4cm]{GER.png} & Germany & \includegraphics[width=0.4cm]{GER.png} & Germany \\ 
  3 & \includegraphics[width=0.4cm]{GER.png} & Germany & \includegraphics[width=0.4cm]{FRA.png} & France & \includegraphics[width=0.4cm]{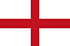} & England \\ 
  4 & \includegraphics[width=0.4cm]{ENG.png} & England & \includegraphics[width=0.4cm]{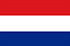} & Netherlands & \includegraphics[width=0.4cm]{FRA.png} & France \\ 
  5 & \includegraphics[width=0.4cm]{NED.png} & Netherlands & \includegraphics[width=0.4cm]{ENG.png} & England & \includegraphics[width=0.4cm]{CAN.png} & Canada \\ 
  6 & \includegraphics[width=0.4cm]{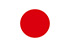} & Japan & \includegraphics[width=0.4cm]{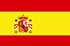} & Spain & \includegraphics[width=0.4cm]{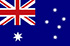} & Australia \\ 
  7 & \includegraphics[width=0.4cm]{AUS.png} & Australia & \includegraphics[width=0.4cm]{CAN.png} & Canada & \includegraphics[width=0.4cm]{JPN.png} & Japan \\ 
  8 & \includegraphics[width=0.4cm]{ESP.png} & Spain & \includegraphics[width=0.4cm]{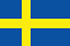} & Sweden & \includegraphics[width=0.4cm]{NED.png} & Netherlands \\ 
  9 & \includegraphics[width=0.4cm]{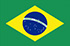} & Brazil & \includegraphics[width=0.4cm]{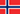} & Norway & \includegraphics[width=0.4cm]{SWE.png} & Sweden \\ 
  10 & \includegraphics[width=0.4cm]{SWE.png} & Sweden & \includegraphics[width=0.4cm]{JPN.png} & Japan & \includegraphics[width=0.4cm]{BRA.png} & Brazil \\ 
  11 & \includegraphics[width=0.4cm]{CAN.png} & Canada & \includegraphics[width=0.4cm]{BRA.png} & Brazil & \includegraphics[width=0.4cm]{NOR.png} & Norway \\ 
  12 & \includegraphics[width=0.4cm]{NOR.png} & Norway & \includegraphics[width=0.4cm]{AUS.png} & Australia & \includegraphics[width=0.4cm]{ESP.png} & Spain \\ 
  13 & \includegraphics[width=0.4cm]{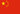} & China PR & \includegraphics[width=0.4cm]{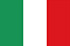} & Italy & \includegraphics[width=0.4cm]{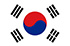} & South Korea \\ 
  14 & \includegraphics[width=0.4cm]{ITA.png} & Italy & \includegraphics[width=0.4cm]{CHN.png} & China PR & \includegraphics[width=0.4cm]{ITA.png} & Italy \\ 
  15 & \includegraphics[width=0.4cm]{KOR.png} & South Korea & \includegraphics[width=0.4cm]{KOR.png} & South Korea & \includegraphics[width=0.4cm]{CHN.png} & China PR \\ 
  16 & \includegraphics[width=0.4cm]{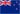} & New Zealand & \includegraphics[width=0.4cm]{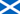} & Scotland & \includegraphics[width=0.4cm]{NZL.png} & New Zealand \\ 
  17 & \includegraphics[width=0.4cm]{SCO.png} & Scotland & \includegraphics[width=0.4cm]{NZL.png} & New Zealand & \includegraphics[width=0.4cm]{SCO.png} & Scotland \\ 
  18 & \includegraphics[width=0.4cm]{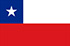} & Chile & \includegraphics[width=0.4cm]{NGA.png} & Nigeria & \includegraphics[width=0.4cm]{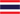} & Thailand \\ 
  19 & \includegraphics[width=0.4cm]{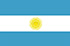} & Argentina & \includegraphics[width=0.4cm]{CHI.png} & Chile & \includegraphics[width=0.4cm]{ARG.png} & Argentina \\ 
  20 & \includegraphics[width=0.4cm]{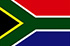} & South Africa & \includegraphics[width=0.4cm]{RSA.png} & South Africa & \includegraphics[width=0.4cm]{NGA.png} & Nigeria \\ 
  21 & \includegraphics[width=0.4cm]{NGA.png} & Nigeria & \includegraphics[width=0.4cm]{THA.png} & Thailand & \includegraphics[width=0.4cm]{CHI.png} & Chile \\ 
  22 & \includegraphics[width=0.4cm]{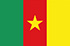} & Cameroon & \includegraphics[width=0.4cm]{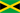} & Jamaica & \includegraphics[width=0.4cm]{CMR.png} & Cameroon \\ 
  23 & \includegraphics[width=0.4cm]{THA.png} & Thailand & \includegraphics[width=0.4cm]{CMR.png} & Cameroon & \includegraphics[width=0.4cm]{RSA.png} & South Africa \\ 
  24 & \includegraphics[width=0.4cm]{JAM.png} & Jamaica & \includegraphics[width=0.4cm]{ARG.png} & Argentina & \includegraphics[width=0.4cm]{JAM.png} & Jamaica \\ 
\end{tabular}
\end{table}

\subsection{Probabilities for FIFA World Cup 2019 Winner} \label{sec:simul}
In this section, the hybrid random forest is applied to (new) data for the 
World Cup 2019 in France (in advance of the tournament) to predict winning probabilities 
for all teams and to predict the tournament course.

The Poisson abilities were estimated by a bivariate Poisson model with a half period of 500 days. 
All matches of the 167 national teams played since 2011-06-01 up to 2019-06-01 are 
used for the estimation, what results in a total of 3418 matches. 
All further predictor variables are taken as the latest values shortly before the World 
Cup (and using the finally announced squads of 23 players for all nations).
The bookmaker consensus abilities are based on the average odds of 18 bookmakers.

For each match in the World Cup 2019, the hybrid random forest can 
be used to predict an expected number of goals for both teams. Given the expected 
number of goals, a real result is drawn by assuming two (conditionally) independent 
Poisson distributions for both scores. Based on these results, all 36 matches from the 
group stage can be simulated and final group standings can be calculated. Due to 
the fact that real results are simulated, we can precisely follow the official FIFA rules when
determining the final group standings\footnote{The final group standings are determined by (1) the number of
points, (2) the goal difference and (3) the number of scored goals.
If several teams coincide with respect to all of these three criteria, a
separate table is calculated based on the matches between the coinciding
teams only. Here, again the final standing of the teams is
determined following criteria (1)--(3). If still no distinct decision can
be taken, the decision is induced by lot.\label{fifa:rules}}. This enables us to 
determine the matches in the round-of-sixteen and we can continue by 
simulating the knockout stage. In the case of draws in the knockout stage, 
we simulate extra-time by a second simulated result. However, here we multiply 
the expected number of goals by the factor 0.33 to account for the shorter time 
to score (30 min instead of 90 min). In  the case of a further draw in extra-time 
we simulate the penalty shootout by a (virtual) coin flip.

Following this strategy, a whole tournament run can be simulated, which we repeat 100,000
times. Based on these simulations, for each of the 24 participating
teams probabilities to reach the single knockout stages and,
finally, to win the tournament are obtained. These are summarized
in Table~\ref{winner_probs} together with the (average) winning probabilities
based on 18 different bookmakers for comparison.

\begin{table}[!h]
\small
\caption{\label{winner_probs}Estimated probabilities (in \%) for reaching the 
different stages in the FIFA World Cup 2019 for all 24 teams based on 100,000 
simulation runs of the FIFA World Cup together with (average) winning probabilities 
based on the odds of 18 bookmakers.}\vspace{0.4cm}

\centering
\begin{tabular}{lllrrrrrr}
   \toprule
 &  &  & Round & Quarter & Semi & Final & World & Bookmakers \\ 
   &  &  & of 16 & finals & finals &  & Champion &  \\ 
   \midrule
1. & \includegraphics[width=0.4cm]{USA.png} & USA & 98.4 & 75.5 & 53.4 & 39.6 & 28.1 & 17.7 \\ 
  2. & \includegraphics[width=0.4cm]{FRA.png} & FRA & 95.9 & 66.8 & 40.7 & 25.4 & 14.3 & 18.2 \\ 
  3. & \includegraphics[width=0.4cm]{ENG.png} & ENG & 96.1 & 69.8 & 45.3 & 23.8 & 13.3 & 11.0 \\ 
  4. & \includegraphics[width=0.4cm]{GER.png} & GER & 95.4 & 66.3 & 36.9 & 22.9 & 12.9 & 12.4 \\ 
  5. & \includegraphics[width=0.4cm]{NED.png} & NED & 92.7 & 47.1 & 25.9 & 12.0 & 5.1 & 6.0 \\ 
  6. & \includegraphics[width=0.4cm]{SWE.png} & SWE & 91.2 & 50.7 & 24.8 & 12.1 & 4.4 & 3.3 \\ 
  7. & \includegraphics[width=0.4cm]{BRA.png} & BRA & 88.7 & 51.2 & 25.5 & 10.5 & 3.9 & 3.8 \\ 
  8. & \includegraphics[width=0.4cm]{AUS.png} & AUS & 89.0 & 50.0 & 24.2 & 10.1 & 3.8 & 4.7 \\ 
  9. & \includegraphics[width=0.4cm]{ESP.png} & ESP & 81.5 & 43.8 & 20.1 & 9.4 & 3.6 & 3.6 \\ 
  10. & \includegraphics[width=0.4cm]{JPN.png} & JPN & 82.5 & 43.3 & 21.1 & 8.0 & 2.7 & 5.3 \\ 
  11. & \includegraphics[width=0.4cm]{CAN.png} & CAN & 85.4 & 33.2 & 14.7 & 5.7 & 2.0 & 3.1 \\ 
  12. & \includegraphics[width=0.4cm]{ITA.png} & ITA & 81.7 & 38.8 & 16.7 & 5.8 & 1.9 & 1.6 \\ 
  13. & \includegraphics[width=0.4cm]{NOR.png} & NOR & 75.0 & 33.7 & 13.1 & 4.6 & 1.5 & 2.2 \\ 
  14. & \includegraphics[width=0.4cm]{CHN.png} & CHN & 72.5 & 29.0 & 9.5 & 3.1 & 0.8 & 1.5 \\ 
  15. & \includegraphics[width=0.4cm]{SCO.png} & SCO & 66.6 & 24.5 & 8.3 & 2.4 & 0.7 & 0.9 \\ 
  16. & \includegraphics[width=0.4cm]{KOR.png} & KOR & 64.8 & 23.6 & 7.3 & 2.0 & 0.5 & 1.2 \\ 
  17. & \includegraphics[width=0.4cm]{NZL.png} & NZL & 65.4 & 16.1 & 4.9 & 1.2 & 0.3 & 1.1 \\ 
  18. & \includegraphics[width=0.4cm]{THA.png} & THA & 36.9 & 7.9 & 1.8 & 0.3 & 0.1 & 0.2 \\ 
  19. & \includegraphics[width=0.4cm]{NGA.png} & NGA & 30.1 & 6.5 & 1.3 & 0.2 & 0.0 & 0.4 \\ 
  20. & \includegraphics[width=0.4cm]{ARG.png} & ARG & 22.6 & 5.2 & 1.0 & 0.2 & 0.0 & 0.7 \\ 
  21. & \includegraphics[width=0.4cm]{CHI.png} & CHI & 26.2 & 5.4 & 1.1 & 0.2 & 0.0 & 0.7 \\ 
  22. & \includegraphics[width=0.4cm]{CMR.png} & CMR & 26.6 & 5.1 & 1.1 & 0.2 & 0.0 & 0.2 \\ 
  23. & \includegraphics[width=0.4cm]{RSA.png} & RSA & 19.6 & 3.9 & 0.8 & 0.1 & 0.0 & 0.3 \\ 
  24. & \includegraphics[width=0.4cm]{JAM.png} & JAM & 15.1 & 2.7 & 0.4 & 0.1 & 0.0 & 0.1 \\ 
   \bottomrule
\end{tabular}

\end{table}

We can see that, according to our hybrid random forest model, 
USA is the favored team with a predicted winning probability of $28.1\%$ 
followed by France, England and Germany. Overall, this result seems in line with the probabilities 
from the bookmakers, as we can see in the last column. However, while the bookmakers slightly favor France, 
the random forest model predicts a clear advantage for USA. 
Beside the probabilities of becoming world champion, Table~\ref{winner_probs} provides some 
further interesting insights also for the single stages within the tournament. For example, it is interesting 
to see that the two favored teams USA and France have similar chances to at least reach the 
round-of-sixteen ($98.4\%$ and $95.9\%$, respectively), while the probabilities to at least reach 
the quarter finals differ significantly. While USA has a probability of $75.5\%$ to reach at least the quarter finals 
and of $53.4$ to reach at least the semi finals, France only achieves respective probabilities of 
$66.8\%$ and $40.7$. Obviously, in contrast to USA, France has 
a rather high chance to meet a strong opponent in the round-of-sixteen and the quarter finals. 

\section{Concluding remarks}\label{sec:conclusion}

In this work, we proposed a  hybrid modeling approach for the scores of
international soccer matches which combines random forests with two different ranking methods,
a Poisson ranking method and abilities based on bookmakers' odds.
While the random forest is principally based on the competing teams' covariate information,
the latter two components provide ability parameters, which serve as adequate 
estimates of the current team strengths a well as of the information contained in the bookmakers' odds. 
In order to combine the methods, the Poisson ranking method needs to be repeatedly applied 
to historical match data preceding each World Cup from the training data. This way, for each 
World Cup in the training data and each participating team current ability estimates are obtained. 
Similarly, the bookmaker consesus abilities are obtained by inverse tournament simulation
based on the aggregated winning odds from several online bookmakers.
These two ability estimates can be added as additional covariates to the set of covariates used 
in the random forest procedure.

Additionally, based on the estimates of the combined ranking-based random forest 
on the training data, we repeatedly simulated the FIFA Women's World Cup 2019 100,000 times. 
According to these simulations, the defending champion USA is the top 
favorite for winning the title, followed by France, England and Germany.
Furthermore, survival probabilities for all teams and at all tournament stages are
provided.

\subsection*{Acknowledgment}
We thank Jonas Heiner for his tremendous effort in helping us to collect
the covariate data.

\bibliographystyle{chicago}
\bibliography{literatur}

\end{document}